\documentclass{article}

\PassOptionsToPackage{numbers, compress}{natbib}

\usepackage[preprint]{neurips_2024}

\usepackage[utf8]{inputenc} 
\usepackage[T1]{fontenc}    
\usepackage{url}            
\usepackage{booktabs}       
\usepackage{amsfonts}       
\usepackage{nicefrac}       
\usepackage{microtype}      
\usepackage{xcolor}         

\usepackage{amsmath,amsfonts,bm}

\def\eqref#1{equation~\ref{#1}}

\def\1{\bm{1}}

\DeclareMathAlphabet{\mathsfit}{\encodingdefault}{\sfdefault}{m}{sl}
\SetMathAlphabet{\mathsfit}{bold}{\encodingdefault}{\sfdefault}{bx}{n}

\definecolor{Gray}{gray}{0.9}
\definecolor{citecolor}{HTML}{2980b9}
\definecolor{linkcolor}{HTML}{c0392b}

\usepackage[pagebackref=true,breaklinks=true,colorlinks,bookmarks=false,citecolor=citecolor,linkcolor=linkcolor]{hyperref}

\usepackage{url}
\usepackage{graphicx}
\usepackage{enumitem}
\usepackage{booktabs}  

\usepackage{multirow}
\usepackage{wrapfig}

\definecolor{Gray}{gray}{0.9}

\newcommand{\names}{\textsc{MultiIoT}}
\newcommand{\iotlm}{\textsc{IoT-LM}}
\newcommand{\iot}{\textsc{IoT}}

\title{\iotlm: Large Multisensory Language Models\\for the Internet of Things}

\author{%
  Shentong Mo, Ruslan Salakhutdinov, Louis-Philippe Morency, Paul Pu Liang\\
  Carnegie Mellon University\\
  \texttt{shentongmo@gmail.com}
}

\begin{document}

\maketitle

\vspace{-4mm}
\begin{abstract}
The Internet of Things (\iot) network integrating billions of smart physical devices embedded with sensors, software, and communication technologies is a critical and rapidly expanding component of our modern world. 
The \iot\ ecosystem provides a rich source of real-world modalities such as motion, thermal, geolocation, imaging, depth, sensors, and audio to recognize the states of humans and physical objects.
Data-driven tools present a rich opportunity to automatically process \iot\ data at scale, enabling efficient inference for understanding human wellbeing, controlling physical devices, and interconnecting smart cities.
To realize this potential, we introduce \iotlm, an open-source large multisensory language model tailored for the \iot\ ecosystem.
\iotlm\ is enabled by two technical contributions: the first is \names, the most expansive unified \iot\ dataset to date, encompassing over 1.15 million samples from 12 modalities and 8 tasks prepared for multisensory pre-training and instruction-tuning. The second is a new multisensory multitask adapter layer to condition pre-trained large language models on multiple multisensory \iot\ tasks simultaneously, enabling the sharing of information across modalities and tasks for better generalization.
Not only does \iotlm\ yield substantial improvements on 8 supervised \iot\ classification tasks, but it also demonstrates new interactive question-answering, reasoning, and dialog capabilities conditioned on \iot\ sensors.
We release \iotlm's data sources and new multisensory language modeling framework at the repository~\footnote{\url{https://github.com/Multi-IoT/MultiIoT}}.
\end{abstract}

\vspace{-2mm}
\section{Introduction}
\vspace{-2mm}

The digital world is witnessing an unprecedented surge in the realm of the Internet of Things (\iot): the ever-growing system of interlinked devices from individual households to vast industrial complexes~\citep{atzori2010internet}. These devices are often embedded with sensors, software, and communication technologies that can safely and privately analyze the human and physical world~\citep{li2015internet,rose2015internet}. For example, high-fidelity sensors are able to recognize physical activities to inform us of our daily physical wellness~\citep{qi2015survey,yuehong2016internet}; vision, depth, and lidar sensors are able to navigate self-driving cars and connect them with traffic lights for efficient transport management~\citep{javaid2018smart,khayyam2020artificial}; and wifi, depth, camera sensors can detect if the elderly require assistance in hospitals~\citep{ahamed2018applying,kulkarni2014healthcare}.
As a result, there has been substantial interest in building machine learning systems that can efficiently process these \iot\ sensors and make predictions, which has great potential for understanding human wellbeing, controlling physical devices, and interconnecting smart cities~\citep{adi2020machine,ghazal2021iot,sworna2021towards,zantalis2019review}.

However, most existing machine learning approaches in the \iot\ domain have largely focused on supervised models, trained only on a single input sensory modality and for a single output prediction task~\cite{Ahuja2021TouchPose,Riku2022RGBDGaze,Huang2018deep,jia2021llvip,kong2021EyeMU}. We extend the existing ML for \iot\ paradigms in two directions. Firstly, in the input space, we investigate how to learn from multiple heterogeneous and interacting \iot\ sensory modalities simultaneously. This is important since practical \iot\ scenarios often see multiple sensors used for different use cases, each with its own unique information, structures, and noise topologies. Secondly, in the output space, we study how to ground large language models (LLMs) on \iot\ sensors to enable simultaneous prediction over many IoT-related real-world tasks. This can enable us to combine the real-world sensing of \iot\ with the dialog, reasoning, and generalization abilities of large language models.
Together, these result in the development of \iotlm, a large multisensory language model capable of processing many \iot\ sensors for a range of predictive, question-answering, reasoning, and interactive dialog tasks. Our primary contributions in \iotlm\ can be summarized as:
\begin{itemize}[noitemsep,topsep=0pt,nosep,leftmargin=*,parsep=0pt,partopsep=0pt]
    \item \textbf{IoT-Language resource}: To train \iotlm, we collect and publicly release a dataset with 1.15 million \iot\ sensor - natural language paired samples covering 12 real-world sensory modalities and $8$ \iot\ tasks. These sensors and tasks are rooted in practical scenarios such as human health and wellness, physical commonsense, and smart cities.
    \item \textbf{\iotlm\ architecture}: The key innovation in \iotlm's architecture is a new multisensory multitask adapter layer to condition pretrained LLMs on multiple multisensory \iot\ tasks simultaneously, enabling the sharing of information across modalities and tasks for better generalization.
    \item \textbf{\iotlm}: With this new \iot\ training data and multisensory multitask adapter, we train \iotlm, the first large multisensory language model that can perceive physical \iot\ sensors. Through sensor-language pretraining and instruction tuning, \iotlm\ displays interactive question-answering, reasoning, and dialog capabilities conditioned on \iot\ sensor data. We publicly release a set of \iotlm\ models spanning 7B to 70B parameters, along with all the curated \iot-language resources, and training code.
\end{itemize}

\vspace{-2mm}
\section{Related Work}
\vspace{-2mm}

We cover related work in the design and applications of \iot\ sensors, how machine learning can be used to accelerate \iot\ perception, and related background work in multisensory machine learning and foundation models.

\textbf{Internet of Things (\iot):} 
The pursuit of extracting meaningful insights from \iot\ data~\cite{atmoko2017IoT,khan2018real,kumar2020time,Cook2020anomaly} has led to various innovative approaches that focus on individual modalities and specific tasks~\cite{Gardner1985exponential,Baki2006exponential,Alysha2011forecasting} and resource-constrained devices~\cite{Ebrahimi2019post,khor2021public,Imteaj2022a}. For instance, DIP-IMU~\cite{Huang2018deep} effectively fuses depth sensing with IMU data for enhanced pose estimation, demonstrating the potential of multimodal integration. Similarly, EyeMU~\cite{kong2021EyeMU} utilizes time-series neural networks to process IMU sensor data for accurate gaze tracking on mobile devices. TouchPose~\cite{Ahuja2021TouchPose} explores tactile data for pose recognition, while LLVIP~\cite{jia2021llvip} focuses on leveraging visual data from \iot\ environments for dynamic real-world applications. Furthermore, RGBDGaze~\cite{Riku2022RGBDGaze} integrates RGBD data for gaze estimation, highlighting the diversity of sensory data applications. However, these approaches generally remain confined to single-task models and lack the capability to generalize across multiple \iot\ modalities and tasks, an area where our work with \iotlm\ introduces a significant advancement by enabling statistical sharing and generalization across a broad spectrum of \iot\ data.

\textbf{Multisensory machine learning:} The field of multimodal machine learning has seen substantial growth, with models designed to integrate inputs from various sensory channels to perform more complex interpretation and interaction with the environment~\citep{baltruvsaitis2017multimodal,liang2022foundations}. Notable works include multimodal transformers~\cite{tsai2019multimodal,lu2019vilbert} that fuse visual, auditory, and textual data to improve learning efficacy. These multimodal transformers have also been scaled for self-supervised pre-training across an increasing range of modalities including time-series and sensors~\cite{jaegle2021perceiverio,liang2022highmmt,reed2022generalist}. While these models offer a foundation for integrating diverse data types, they often do not address the unique challenges posed by \iot\ environments, such as the integration of non-traditional sensor data (e.g., IMU, thermal sensors) and the need for models to interact dynamically with physical environments.

\textbf{Foundation models:} 
Recently, the concept of foundation models~\cite{bommasani2022opportunities,chowdhery2022palm,anil2023palm}, pre-trained on large-scale datasets and adaptable to a wide range of tasks~\cite{,zhu2023minigpt,jiang2023mistral,geminiteam2024gemini}, has gained traction. These models, exemplified by works such as GPT-4~\cite{openai2023gpt4} and LLaMA~\cite{touvron2023llama}, offer a robust starting point for further fine-tuning on individual natural-language tasks. There has also been a recent drive towards large multimodal models, using either LLMs as a starting point and training adapters from other modalities to LLM input space~\cite{gao2023llamaadapterv2,zhu2023minigpt}, or training multimodal transformers from scratch (sometimes called `natively') with interleaved language tokens, image frames, audio frames, and other modalities~\cite{geminiteam2024gemini,openai2023gpt4}. Our approach extends these paradigms into the \iot\ domain, where \iotlm\ adapts a large language model framework to not only understand textual information but also effectively process and reason about multisensory data from diverse \iot\ sensors. This adaptation enables unprecedented capabilities in performing complex reasoning, dialogue, and interactive question-answering tasks directly related to physical \iot\ contexts, setting a new benchmark for intelligent \iot\ systems.

\begin{figure}[t]
\centering
\vspace{-0mm}
\includegraphics[width=0.9\linewidth]{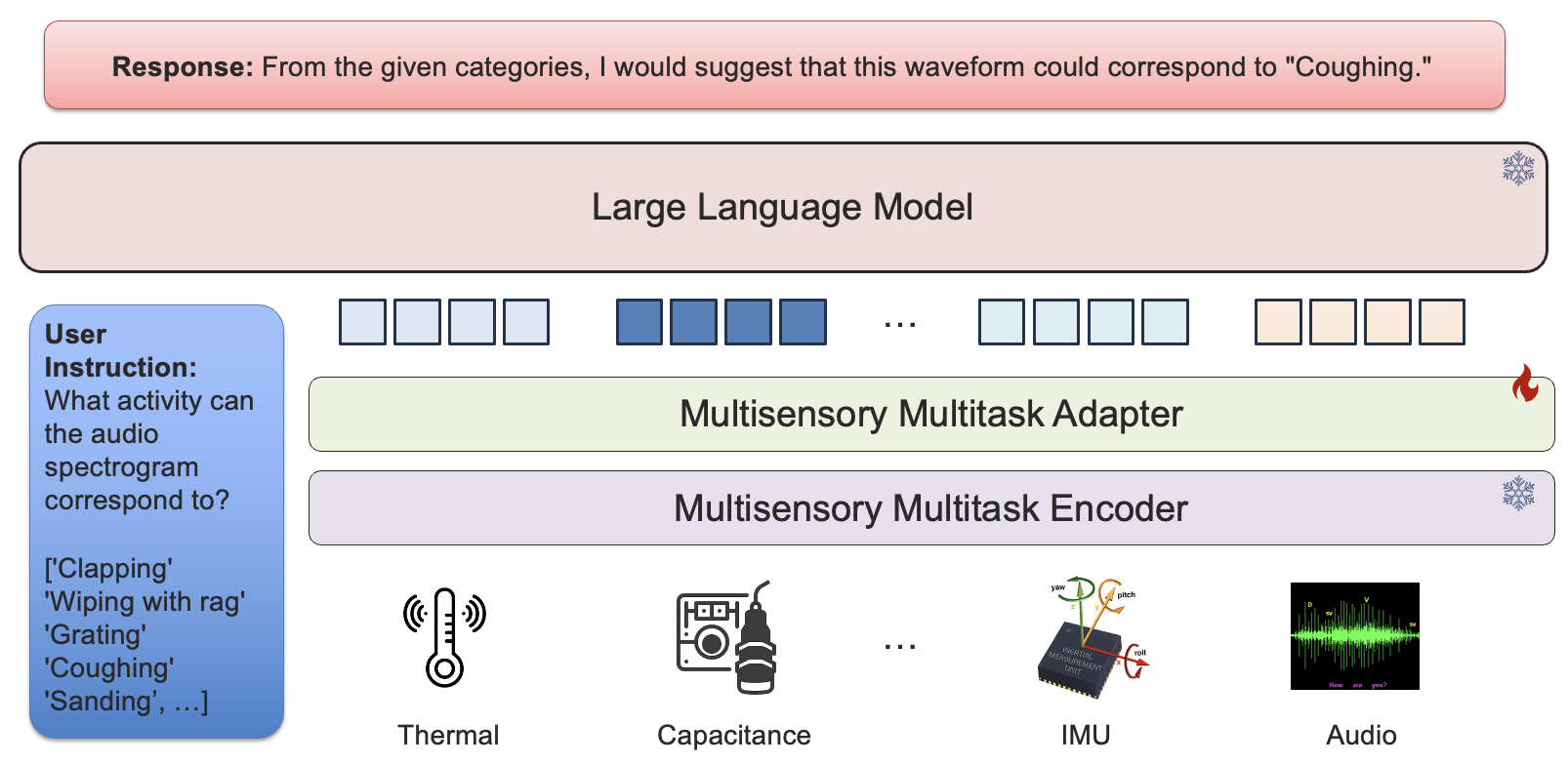}
\vspace{-2mm}
\caption{Illustration of \iotlm\ architecture, highlights the integration of multisensory data through modality-specific encoders and the novel multisensory multitask adapter layer. We illustrate how different sensory inputs are processed, combined, and utilized to adapt a pre-trained language model for \iot\ applications to handle and interpret complex, real-world sensor data efficiently.}
\vspace{-2mm}
\label{fig: main_img_iotlm}
\end{figure}

\vspace{-2mm}
\section{\iotlm: A New Multisensory \iot\ Foundation Model}
\vspace{-2mm}

In this section, we describe the architecture, data curation, and training innovations in \iotlm.

\vspace{-1mm}
\subsection{\iotlm\ Architecture: Multisensory Multitask Adapter}
\vspace{-1mm}

\iotlm's backbone consists of a variety of \iot\ sensory inputs, a general-purpose multisensory multitask encoder that fuses the information from multiple sensors and across multiple tasks, a multisensory multitask adapter that transforms encoded representations into pretrained LLM input space, and the pretrained LLM itself. We show an overview of the \iotlm\ architecture in Figure~\ref{fig: main_img_iotlm}.

\paragraph{Multisensory multitask encoder for heterogeneous \iot\ signals} 
The encoder is trained using a novel approach that effectively manages the heterogeneity of IoT data by combining supervised learning with unsupervised feature extraction techniques. 
This training involves task-specific adaptations, where the encoder learns to map raw sensor data to an intermediate feature space conducive for language model processing. The encoder handles the multisensory data through a combination of multimodal fusion methods, including early, late, and model-based fusion, which are chosen based on the data characteristics and the specific requirements of each task.

Specifically, the adapters transform the sensor data into a representation that is more conducive for the language model to process. For each sensor modality \(x_i\), we employ a dedicated encoder \(E_i\) that maps raw sensor data to an intermediate feature space. The encoded features from different modalities are then combined using a fusion mechanism within the adapter, allowing the model to harness information from all available sensors. 
We employ late fusion methods that are extracted from each sensor modality and are independently processed and only combined at decision-making layers, facilitating specialization in feature extraction while leveraging multimodal data for final predictions.
These fusion techniques are selected and fine-tuned based on the specific characteristics of the data and the tasks at hand, ensuring optimal performance. 
Our fusion methods innovations involve developing new model-based fusion techniques that adaptively adjust the weighting of different sensor inputs based on their predictive value for specific tasks, a novel approach to handling IoT data heterogeneity.
The training of these components is conducted through a multitask supervised learning framework that emphasizes individual task accuracy and enhances generalizability across tasks by sharing representations and leveraging common patterns found in the diverse IoT data landscape. 
This method strengthens the model's ability to perform well across various conditions and tasks, making it highly effective for real-world IoT applications.

\paragraph{Multisensory multitask adapter layer} We extend the typical use of adapter modules, which are compact, trainable layers inserted between the existing layers of a pre-trained language model~\cite{touvron2023llama}. These adapters are designed to fine-tune the pre-existing model to new tasks and modalities without significant modifications to the original model's weights, thus preserving its general linguistic capabilities while extending its functionality to new IoT-specific domains.

\begin{figure}[t]
\centering
\vspace{-0mm}
\includegraphics[width=0.6\linewidth]{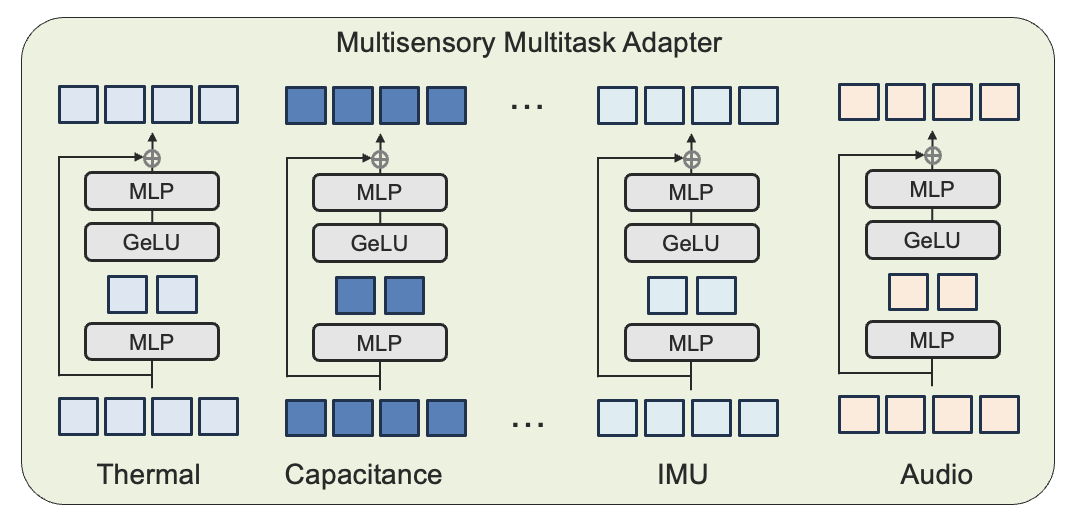}
\vspace{-2mm}
\caption{Illustration of the multisensory multitask adapter layer we designed for \iotlm. The adapter takes multiple sensor features extracted from dedicated input encoders, performs multimodal fusion into higher-order representations, and simultaneously transforms all fused multimodal features into the same representation space for an LLM to process.}
\vspace{-2mm}
\label{fig: main_img_adapter}
\end{figure}

The key difference in \iotlm\ is that the multisensory multitask adapter layer conditions pretrained LLMs on multiple multisensory \iot\ tasks simultaneously, enabling the sharing of information across modalities and tasks for better generalization and holistic understanding of \iot\ environments. We show an overview of the multisensory multitask adapter layer in Figure~\ref{fig: main_img_adapter}. The architecture is formalized as follows:
\begin{equation}
    y = M_{W+A}(E_1(x_1) \oplus E_2(x_2) \oplus ... \oplus E_m(x_m)) = M_W(A_{W_A}(E_1(x_1) \oplus E_2(x_2) \oplus ... \oplus E_m(x_m)))
\end{equation}
where \(M_{W+A}(\cdot)\) and \(M_W(\cdot)\) denote the models with combined and original weights, respectively, and \(E_1, E_2, ..., E_m\) represent the encoders for each sensor modality. This structure allows \iotlm\ to maintain the foundational knowledge of the pre-trained model while adapting to the specific nuances of \iot\ data.

\vspace{-1mm}
\subsection{\iot-Language Data Collection}
\vspace{-1mm}

To train \iotlm, we aggregate and release the largest \iot\ dataset for machine learning research, comprising 1.15 million samples from 12 different sensory modalities for 8 different real-world \iot\ prediction tasks related to personal wellness, healthcare, smart cities, and more. These modalities include Inertial Measurement Units (IMU), thermal sensors, GPS, LiDAR, gaze, pose, capacitance sensors, and traditional modalities such as images, audio, and video. Tasks supported include:
\begin{enumerate}[noitemsep,topsep=0pt,nosep,leftmargin=*,parsep=0pt,partopsep=0pt]
    \item \textbf{Gaze estimation:} For human-computer interaction, driver monitoring, and virtual reality applications, our dataset includes RGB images of faces, depth data, and IMU outputs. The model is tasked with predicting X/Y coordinates for gaze tracking, necessitating a deep understanding of the interactions between these modalities.
    
    \item \textbf{Depth estimation:} This task involves estimating the distance from cameras to objects in images, vital for AR/VR, robotics, and object detection. Our dataset includes RGB images combined with camera parameters, GPS coordinates, and IMU data to create depth maps for various scenarios, including street scenes and robotic hand interactions.
    
    \item \textbf{Gesture classification:} Key to enhancing human-machine interfaces, this task uses data from gaze tracking and IMU sensors (accelerometer, gyroscope, and orientation) to classify human gestures. The challenge here is to accurately interpret the nuanced cross-modal interactions.
    
    \item \textbf{Pose estimation:} This task determines the arrangement of human joints, using RGB images and IMU data to predict poses of the human body, including 24 joints with three angles each (yaw, pitch, roll). This requires the model to fuse data from IMUs and visual inputs.
    
    \item \textbf{Touch contact:} To improve touch-based device interactions, this task classifies the type of touch on capacitive surfaces using RGB and capacitive images, depth maps, and hand poses.
    
    \item \textbf{Event detection:} In applications ranging from healthcare to smart homes, this task identifies specific occurrences or anomalies in data streams, using audio spectrograms and IMU data to categorize events over time.
    
    \item \textbf{Activity recognition:} Central to applications in fitness and healthcare, this task uses RGB images, pose data, and IMU outputs to recognize human activities such as walking, running, or jumping.
    
    \item \textbf{3D reconstruction:} Significant in gaming, film, and AR/VR, this task involves creating three-dimensional models from RGB images, capacitance images, and depth maps, aimed at reconstructing 3D poses of objects and environments.
\end{enumerate}

These diverse tasks not only prepare \iotlm\ to handle complex sensor data but also ensure it can perform a broad range of functions, from classification to complex reasoning and dialogue involving multiple \iot\ devices.

\vspace{-1mm}
\subsection{Multisensory Multitask Encoder Pretraining}
\vspace{-1mm}

Building on the \iotlm\ architecture and data resources, we now describe the pre-training and instruction tuning stages in \iotlm. The pre-training stage aims to learn a general-purpose multisensory multitask encoder that fuses the information from multiple sensors and across multiple tasks. 
During pretraining, the multisensory multitask encoder is trained on a combination of supervised learning tasks to extract and fuse features from diverse sensor data effectively. This stage is pivotal in preparing the model to process and understand complex sensory inputs, which can be mathematically described as follows:
\begin{equation}
    \Theta_{\text{pre}} = \arg \min_\theta \sum_{k=1}^K\sum_{(x_k, y_k) \in \mathcal{D}_k} \mathcal{L}_k(M_{\theta}(E_k(x_k)), y_k),
\end{equation}
where \(\theta\) represents the parameters of the entire network including the adapters, \(\mathcal{D}_k\) is the dataset consisting of input-output pairs \((x_k, y_k)\), \(E_k(x_k)\) represents the encoded inputs from various sensors for task $k$, and \(\mathcal{L}_k\) denotes the loss function used to measure the discrepancy between the model's predictions and the true outputs.

\begin{figure}[t]
\centering
\vspace{-0mm}
\includegraphics[width=0.95\linewidth]{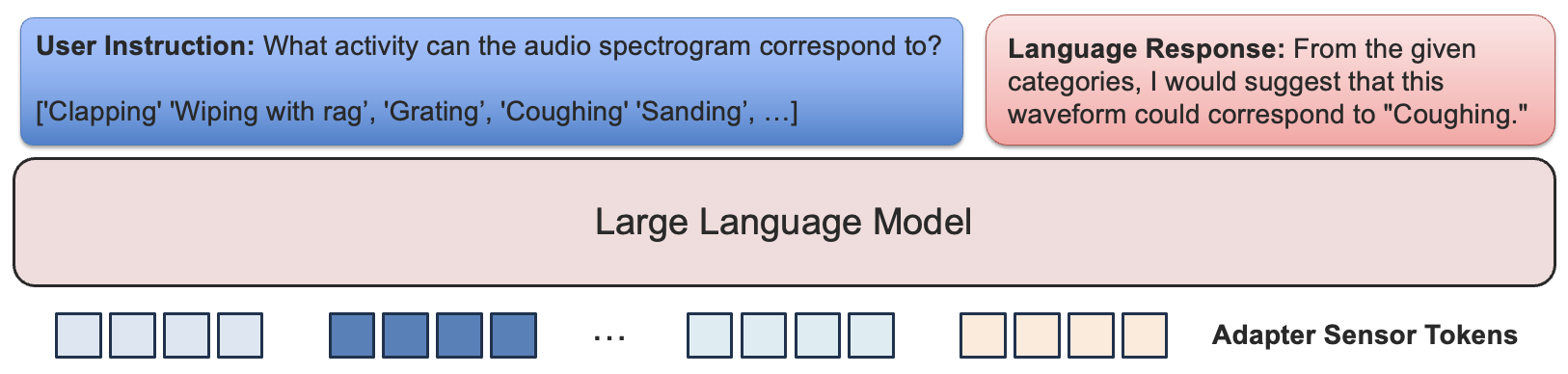}
\vspace{-2mm}
\caption{Illustration of \iotlm\ instruction tuning paradigm that learns to perform specific tasks based on directive inputs. By training on a diverse range of input modalities and output tasks, this enables \iotlm\ to process multiple \iot\ inputs and execute complex tasks.}
\vspace{-2mm}
\label{fig: main_img_pretrain}
\end{figure}

\vspace{-1mm}
\subsection{\iotlm\ Instruction Tuning}
\vspace{-1mm}

During the instruction-tuning phase, \iotlm\ is trained to understand and interpret multisensory \iot\ data contexts and perform specific tasks based on directed language inputs. These tasks can include making a prediction, answering a question, holding a dialog, reasoning about physical properties, and so on. This stage is crucial for refining the model's ability to follow complex and varied human instructions.
The instruction tuning phase is guided by the following optimization:
\begin{equation}
    \Theta_{\text{tune}} = \arg \min_\theta \sum_{k=1}^K\sum_{(x_k, c_k, y_k) \in \mathcal{T}_k} \mathcal{L}_k(M_{\theta}(E_k(x), c_k), y_k),
\end{equation}
where \(\mathcal{T}_k\) represents the task-specific dataset with inputs \(x_k\), context or commands \(c_k\), and outputs \(y_k\) for task $k$. Here, \(E_k(x_k)\) denotes the encoded sensor data for task $k$, \(c_k\) provides specific instructions or task directives, and \(M_{\theta}\) is the model parameterized by \(\theta\) including the tuned adapters.
This phases ensure that \iotlm\ not only learns to process a wide array of sensor data but also understands and executes complex commands pertinent to \iot\ applications, thus achieving high performance across varied \iot\ tasks.

The combination of pretraining with IoT-specific sensory data and subsequent instruction tuning prepares \iotlm\ to handle real-world tasks effectively. This training approach not only enhances the model's capability to understand complex sensor data but also ensures it can perform a wide array of tasks, from simple classification to complex reasoning and dialogues involving multiple \iot\ devices.

\vspace{-2mm}
\section{Experiments}
\vspace{-2mm}

Our experiments aim to benchmark the performance of \iotlm\ on supervised \iot\ classification tasks, as well as their reasoning, dialog, interaction, and zero-shot and few-shot transfer abilities across different modalities and tasks.

\vspace{-1mm}
\subsection{Experimental Setup}
\vspace{-1mm}

Experiments were conducted using NVIDIA A100 GPUs, ensuring high-performance computation for our deep learning models. The models were trained for 30 epochs using the Adam optimizer with a learning rate of 1e-4 and a batch size of 16. We compare to the following baselines:

\begin{itemize}[noitemsep,topsep=0pt,nosep,leftmargin=*,parsep=0pt,partopsep=0pt]
    \item \textbf{Unimodal models}~\cite{kong2021EyeMU,Ahuja2021TouchPose,Mollyn2022samosa} processed data from single sensor types using tailored neural architectures for each \iot\ domain data,

    \item \textbf{Unimodal Adapter models}~\cite{gao2023llamaadapterv2} utilized deep architectures such as LLaMA-adpater~\cite{gao2023llamaadapterv2} with adapter layers, fine-tuning only the adapter modules with a learning rate of 0.0005,

    \item \textbf{Unimodal multi-task models}~\cite{liang2021multibench} employed shared encoder layers and task-specific decoders, ensuring balanced gradients among tasks,

    \item \textbf{Multisensory models}~\cite{jaegle2021perceiver}, where data fusion can occur at varying levels, from input to decision, and models ensured balanced data representation from each modality,

    \item \textbf{Multisensory multitask models}~\cite{liang2022highmmt} utilized modality-specific encoders followed by task-specific decoders, balancing both modalities and tasks during training. Each method's efficacy was validated on respective datasets.
\end{itemize}

To evaluate performance, we employ task-specific metrics.  For gaze and pose estimation, we measure the mean euclidean error in centimeters between predictions and ground truth. Depth estimation utilizes mean absolute error in millimeters, while gesture classification, touch contact classification, and activity recognition rely on accuracy metrics. Event detection employs the F1 score for confident threshold predictions, and 3D pose reconstruction is assessed using the End-point-error in millimeters for joint discrepancies.

\begin{table}[t]
\centering
\caption{Multisensory multi-task learning is a particularly effective approach on \iotlm, enabling information sharing to learn general representations for IoT data. 
Our \iotlm\ achieves the best results across all diverse tasks, benefiting from multisensory multitask adapter and instruction tuning.
}
\label{tab: exp_sota}
\scalebox{0.67}{
\begin{tabular}{lcccccccc}
    \toprule
    \multirow{2}{*}{Method} & Gaze est. & Depth est. & Gesture cls. & Pose est. & Touch cls. & Event det. & Activity recog. & 3D recons. \\
    & (cm, $\downarrow$) & (mm, $\downarrow$) & (\%,$\uparrow$) & (cm, $\downarrow$) & (\%, $\uparrow$) & (\%, $\uparrow$) & (\%, $\uparrow$) & (mm, $\downarrow$) \\
    \midrule
    Domain-specific  & 3.76 & 40.9 & 68.2 & 10.86  & 52.6 & 59.3 & 48.5 & 35.6 \\
    Unimodal & 2.26 & 20.7 & 97.3 & 6.49	  & 88.0 & 86.9 & 79.2 & 22.2 \\
    Unimodal Adapter & 2.05 & 18.6 & 97.6 & 5.75	  & 88.7 & 87.5 & 82.3 & 21.3 \\
    Unimodal Multi-task & 1.95 & 18.2 & 98.2 & 5.36 & 89.3 & 88.1 & 82.5 & 20.5 \\
    Multisensory & 1.79 & 17.3 & 98.7 & 4.62	  & 91.2 & 89.1 & 83.5 & 19.6 \\
    Multisensory Multi-task & 1.08 & 13.6 & 99.3 & 3.85 & 93.8 & 92.7 & 87.5 & 17.5 \\
    Multisensory multitask \iotlm & \bf 0.95 & \bf 11.5 & \bf 99.6 & \bf 3.24 & \bf 94.6 & \bf 93.8 & \bf 89.2 & \bf 16.3 \\
    \bottomrule
\end{tabular}}
\vspace{-2mm}
\end{table}

\vspace{-2mm}
\subsection{Main quantitative results}
\vspace{-1mm}

\textbf{Overall performance}: Table~\ref{tab: exp_sota} reports the quantitative results of \iotlm\ compared to state-of-the-art domain-specific, single modality, single task, multimodal multitask, and adapter and alignment models.
As seen in Table~\ref{tab: exp_sota}, the \iotlm\ consistently outperforms the single modality and single task models across all tasks.
This can be attributed to their ability to integrate textual information across modalities and tasks, which is especially crucial when one modality might have noisy or incomplete data.
While the adapter and multimodal multitask models show commendable performance due to their ability to adapt to new tasks, they often fall short in scenarios where multiple modalities have to be processed simultaneously. 
\iotlm\ improves upon models that only perform multimodal multitask supervised learning, since it inherits the prediction and reasoning capabilities from the pretrained large language model component. 

\begin{table}[t]
\centering
\caption{Multimodal learning enables complementary learning of information for \iotlm\ and achieves strong performance. 
For all tasks, the incorporation of more modalities resulted in more robust and accurate models across diverse \iot\ applications based on multiple sensors.
}
\label{tab: exp_modality}
\scalebox{0.75}{
\begin{tabular}{ccccccccc}
    \toprule
    \multirow{2}{*}{Modality Ratio} & Gaze est. & Depth est. & Gesture cls. & Pose est. & Touch cls. & Event det. & Activity recog. & 3D recons. \\
    & (cm, $\downarrow$) & (mm, $\downarrow$) & (\%,$\uparrow$) & (cm, $\downarrow$) & (\%, $\uparrow$) & (\%, $\uparrow$) & (\%, $\uparrow$) & (mm, $\downarrow$) \\
    \midrule
    single-modality  & 1.38 & 15.1 & 98.1 & 5.07 & 91.9 & 91.8 & 86.3 & 18.7 \\
    25\% & 1.25 & 14.3 & 98.4 & 4.63 & 92.5 & 92.2 & 86.7 & 18.3 \\
    50\% & 1.12 & 13.2 & 98.9 & 4.15 & 93.2 & 92.5 & 87.2 & 17.9 \\
    all & \bf 1.08 & \bf 12.9 & \bf 99.2 & \bf 3.76 & \bf 93.5 & \bf 92.9 & \bf 88.1 & \bf 17.5 \\
    \bottomrule
\end{tabular}}
\vspace{-3mm}
\end{table}

\begin{table}[t]
\centering
\caption{Multi-task learning is another effective strategy on \iotlm, enabling information sharing across tasks. 
For all tasks, the incorporation of more tasks resulted in more accurate models.
}
\label{tab: exp_task}
\scalebox{0.75}{
\begin{tabular}{ccccccccc}
    \toprule
    \multirow{2}{*}{Task Ratio} & Gaze est. & Depth est. & Gesture cls. & Pose est. & Touch cls. & Event det. & Activity recog. & 3D recons. \\
    & (cm, $\downarrow$) & (mm, $\downarrow$) & (\%,$\uparrow$) & (cm, $\downarrow$) & (\%, $\uparrow$) & (\%, $\uparrow$) & (\%, $\uparrow$) & (mm, $\downarrow$) \\
    \midrule
    single-task  & 1.38 & 15.1 & 98.1 & 5.07 & 91.9 & 91.8 & 86.3 & 18.7 \\
    25\% & 1.29 & 14.5 & 98.3 & 4.86 & 92.2 & 92.1 & 86.7 & 18.4 \\
    50\% & 1.22 & 13.8 & 98.6 & 4.52 & 92.6 & 92.5 & 87.2 & 18.0 \\
    all & \bf 1.13 & \bf 13.1 & \bf 99.1 & \bf 4.23 & \bf 93.1 & \bf 92.8 & \bf 87.8 & \bf 17.5 \\
    \bottomrule
\end{tabular}}
\end{table}

\textbf{Performance across increasing modalities}: 
In this section, we study how adding additional modalities to \iotlm\ impacts performance. From Table~\ref{tab: exp_modality}, we find significant performance improvements observed when adding multimodal datapoints of increasing ratios (25\%, 50\%, all) as compared to unimodal models.
This can be attributed to the \iotlm's ability to tap into complementary information present in different modalities, especially in scenarios where one modality might be ambiguous or noisy.

\textbf{Performance across increasing tasks}: We also analyzed model performance when trained on increasing numbers of tasks, while keeping the same modality inputs constant. From Table~\ref{tab: exp_task}, we see that \iotlm's performance steadily increased as we increased the number of tasks during training.
This suggests that the multitask instruction tuning in \iotlm\ was beneficial since the model learns more general features while also improving computational efficiency.

\begin{table}[t]
\centering
\caption{\iotlm\ shows the best zero-shot and few-shot generalization capabilities as compared to other supervised unimodal, multimodal, single-task, and multitask variants. As a result, \iotlm\ can be a promising approach to deal with limited labeled data often seen in real-world IoT systems.}
\label{tab: exp_zeroshot}
\scalebox{0.75}{
\begin{tabular}{lcc}
    \toprule
    Method & Gaze estimation (cm, $\downarrow$) & Touch contact classification (\%, $\uparrow$) \\
    \midrule
    multisensory \iotlm\ & 1.08 & 93.5 \\
    multisensory multitask \iotlm\ & \bf 1.03 & \bf 94.1 \\ \hline
    multisensory multitask \iotlm\ (zero-shot) & 1.25 & 92.3 \\
    multisensory multitask \iotlm\ (5-shot) & 1.21 & 92.5 \\
    multisensory multitask \iotlm\ (10-shot) & 1.13 & 92.9 \\
    multisensory multitask \iotlm\ (20-shot) & \bf 1.06 & \bf 93.6 \\
    \bottomrule
\end{tabular}}
\end{table}

\textbf{Zero-shot and few-shot transfer}: 
Furthermore, we study whether \iotlm\ trained on certain modalities or tasks can transfer to a new set of target modalities or tasks they have never seen during training (zero-shot) or have seen with only very few examples (few-shot). 
We chose the fix-8 dataset as the target, primarily because of its diverse representation of modalities (IMU, capacitance, depth, image) and its challenging task (gaze estimation and touch contact classification).
From Table~\ref{tab: exp_zeroshot}, we find that across the board, even a few examples significantly boosted performance compared to the zero-shot setting, which highlights the model's ability to quickly adapt to new information.
Using \iotlm\ as a base model for zero-shot and few-shot experiments consistently outperformed other types of models, such as supervised unimodal, multimodal, single-task, and multitask variants. These gains were most pronounced in the 20-shot setting but were noticeably beneficial even in the 5-shot scenario.
Our results suggest that \iotlm\ excels at zero-shot and few-shot learning not seen in single-modality, single-task, and supervised models, and is a promising approach to deal with limited labeled data often seen in real-world \iot\ systems.

\begin{table}[t]
\centering
\caption{Scaling law of multisensory multi-task adapter is observed on \names, enabling models with more parameters to learn general representations for IoT data. 
}
\label{tab: exp_scaling}
\scalebox{0.75}{
\begin{tabular}{lcccccccc}
    \toprule
    \multirow{2}{*}{Params} & Gaze est. & Depth est. & Gesture cls. & Pose est. & Touch cls. & Event det. & Activity recog. & 3D recons. \\
    & (cm, $\downarrow$) & (mm, $\downarrow$) & (\%,$\uparrow$) & (cm, $\downarrow$) & (\%, $\uparrow$) & (\%, $\uparrow$) & (\%, $\uparrow$) & (mm, $\downarrow$) \\
    \midrule
    7B & 1.03 & 12.7 & 99.4 & 3.56 & 94.1 & 93.1 & 88.3 & 17.1 \\
    13B & 0.98 & 11.3 & 99.5 & 3.42 & 94.3 & 93.3 & 88.5 & 16.6 \\
    70B & \bf 0.95 & \bf 11.5 & \bf 99.6 & \bf 3.24 & \bf 94.6 & \bf 93.8 & \bf 89.2 & \bf 16.3 \\
    \bottomrule
\end{tabular}}
\end{table}

\textbf{Scaling law of \iotlm}: 
Finally, we systematically increased the model size of \iotlm\ to observe the impact of size on performance and representation learning.
We evaluated three configurations of the multisensory multi-task adapter: small (7 billion parameters), medium (13 billion parameters), and large (70 billion parameters), and the results are reported in Table~\ref{tab: exp_scaling}.
Each configuration was trained using a consistent training regime on a curated dataset comprising various \iot\ sensory inputs, including visual, auditory, and tactile data. 
We focused on a range of tasks, such as anomaly detection, predictive maintenance, and activity recognition, to test the adaptability and efficiency of the models at different scales.
Our experiments demonstrate a clear scaling law: as the number of parameters increases, the models exhibit improved performance across all tasks. This is quantified not only in terms of accuracy, but also in how effectively the models generalize to unseen data, indicating better learning of underlying representations.
The observed scaling law suggests that larger models are more adept at integrating and processing multisensory data, leading to more robust and general representations. 
This supports the hypothesis that model capacity plays a crucial role in multisensory learning environments typical of \iot\ applications.

\begin{figure}[t]
\centering
\includegraphics[width=0.98\linewidth]{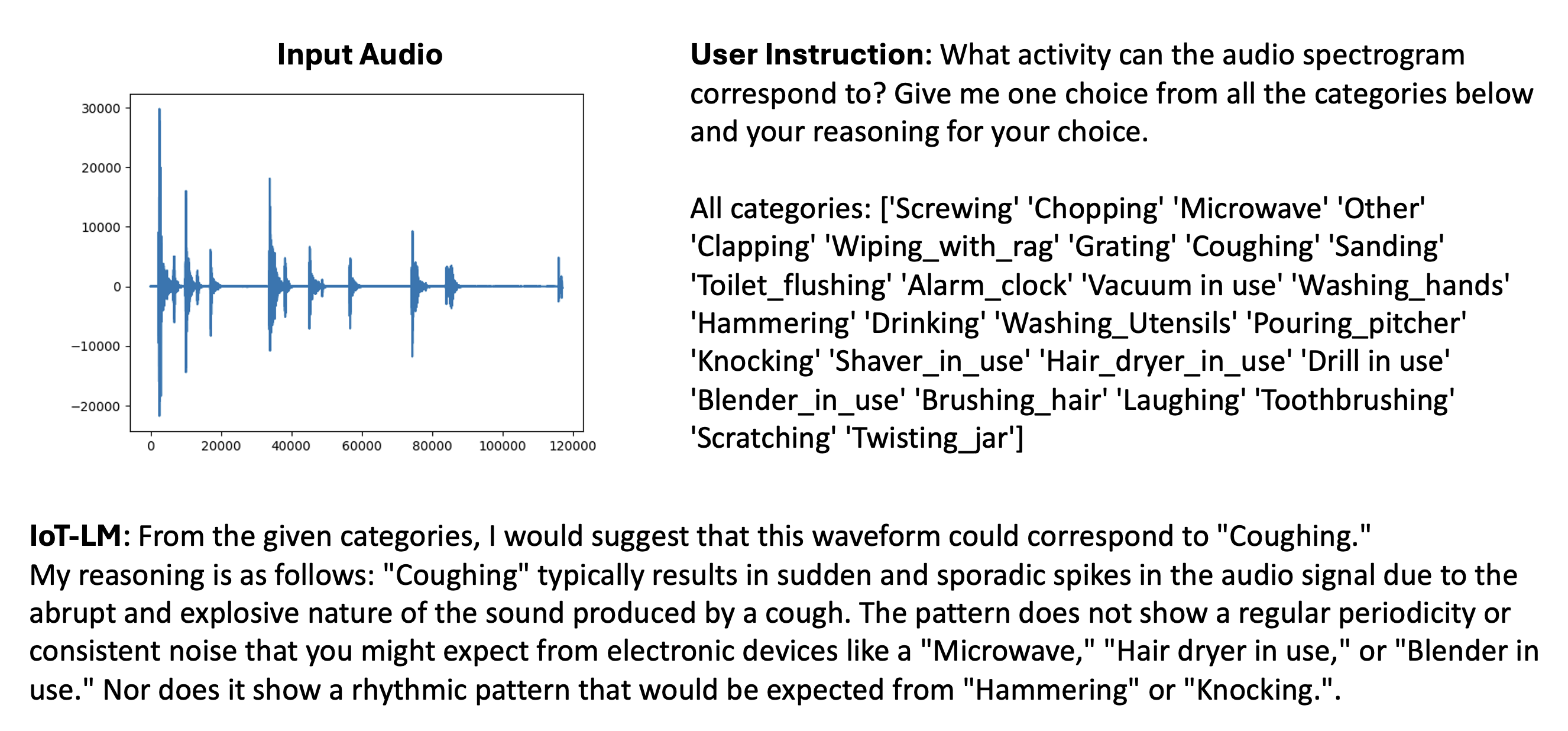}
\vspace{-2mm}
\caption{Dialog for audio example. 
Our \iotlm\ accurately predicts the activity corresponding to the input audio spectrogram, and gives a reasonable explanation for its prediction.}
\vspace{-2mm}
\label{fig: vis_dialog_audio}
\end{figure}

\begin{figure}[t]
\centering
\includegraphics[width=0.98\linewidth]{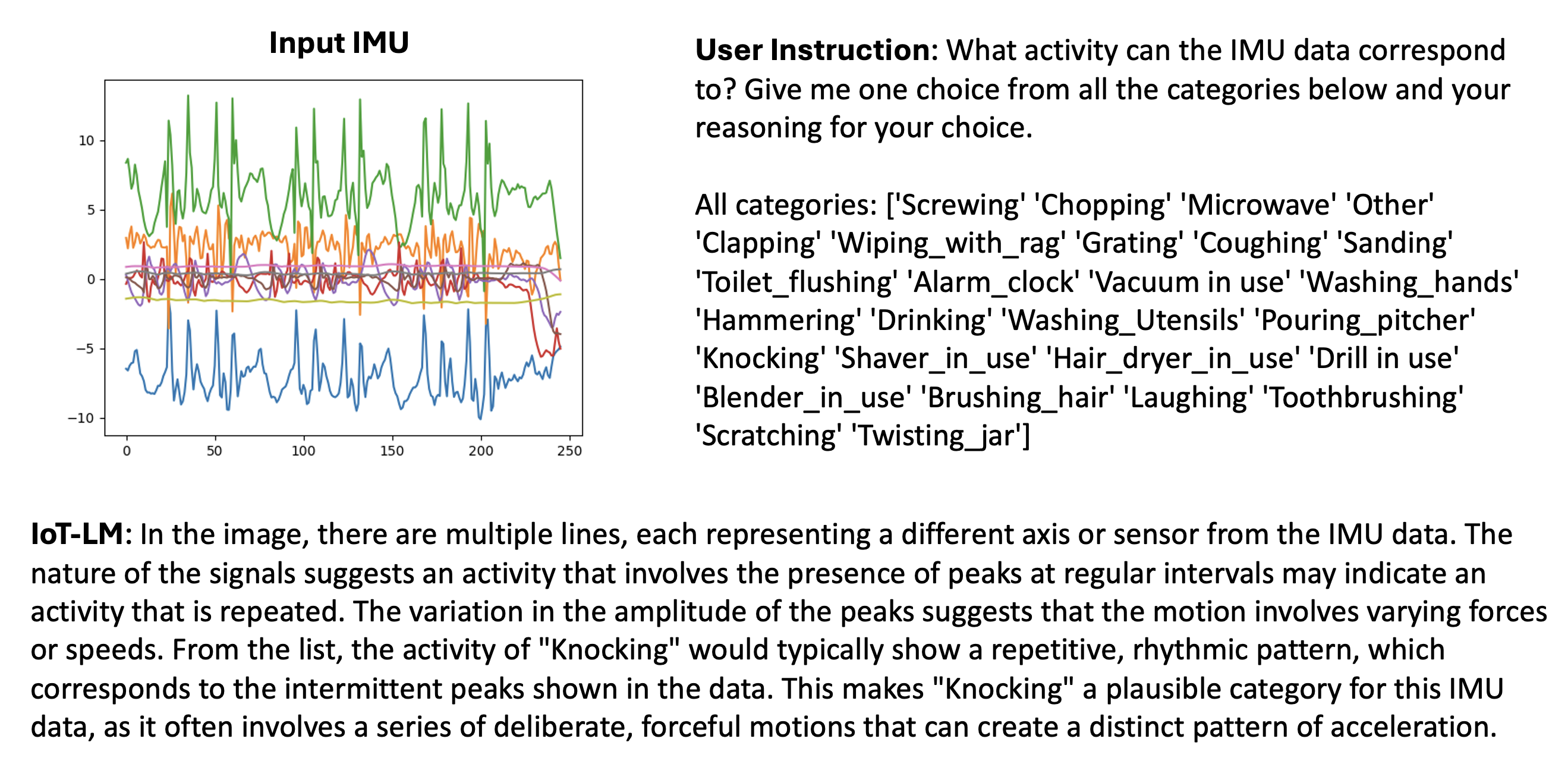}
\vspace{-2mm}
\caption{Dialog for IMU example. 
\iotlm\ accurately predicts the activity corresponding to the input IMU data, and also gives a reasonable explanation for its prediction.}
\vspace{-2mm}
\label{fig: vis_dialog_imu}
\end{figure}

\subsection{Qualitative analysis}

In this section, we qualitatively demonstrate the dialog and interactive capabilities of \iotlm\ on processing audio and Inertial Measurement Unit (IMU) data.

\textbf{Audio}: For audio data, in Figure~\ref{fig: vis_dialog_audio}, \iotlm\ was presented with an audio waveform and instructed to identify the corresponding activity from a list of categories. The model was able to discern that the audio signal most closely resembled the category "Coughing." The choice was justified by the sudden and sporadic nature of the spikes in the waveform, which align with the acoustic signature of a cough. This example highlights the model's capability to process temporal acoustic features and correctly categorize and reason about it.

\textbf{IMU}: For IMU data, in Figure~\ref{fig: vis_dialog_imu}, \iotlm\ analyzed a graph with multiple lines, each representing different sensor readings from the IMU. The task was to select an activity from a list that best matched the IMU data pattern. The model identified "Knocking" as the most likely activity, reasoning that the regular intervals of peaks were indicative of a repetitive action with varying force, which is consistent with the pattern of knocking.

These examples demonstrate the multisensory multi-task adapter's analytical proficiency in interpreting and classifying data from different \iot\ modalities based on their temporal characteristics. 
Such capability is instrumental in realizing the full potential of \iot\ systems, where understanding and acting upon such heterogeneous data in real time is essential for various applications, from smart homes to industrial automation. The examples provided also show the model's potential in bridging the gap between raw sensor data and meaningful insights, enabling non-expert users to interact with and benefit from complex \iot\ systems.

\vspace{-2mm}
\section{Conclusion and Broader Impacts}
\vspace{-2mm}

This paper presents \iotlm, a new large multisensory language model with multisensory perception and natural language interaction capabilities over a spectrum of \iot\ modalities and applications. Key innovations in \iotlm\ include a new multisensory multitask adapter to simultaneously condition pretrained LLMs on multiple multisensory \iot\ tasks for better generalization, as well as a new resource of 1.15 million \iot\ sensor - natural language paired samples covering 12 modalities and 8 real-world tasks. Overall, \iotlm\ not only advances the state-of-the-art in \iot\ predictive learning but also enables new question-answering, reasoning, and interactive dialog capabilities on physical sensor data. Future work should focus on expanding the model’s capabilities to more \iot\ modalities and tasks, improving its reasoning and robustness, and exploring the implications of its deployment in real-world \iot\ systems. We hope that \iotlm\ will inspire further innovations at the intersection of machine learning and \iot, contributing to smarter and more responsive technologies that can understand and interact with their environments.

We are aware of some potential \textbf{limitations and broader impacts} of our work. Firstly, there may be privacy risks associated with making predictions from multimodal data of recorded human behaviors, such as video, audio, activities, poses, and wearable sensors. As such, we made sure the datasets we use are those that collect data from consenting participants. We only use these datasets for research purposes. All data was anonymized and stripped of all personal (e.g., personally identifiable information) and protected attributes (e.g., race, gender). Furthermore, it is also important to keep data and features private on each device without sending it to other locations. There are potential avenues to combine the multisensory and multitask models in our paper with techniques such as federated learning~\citep{DBLP:journals/corr/abs-1812-06127,liang2020think}, differential privacy~\citep{geyer2017differentially}, or encryption~\citep{dankar2013practicing} to preserve the privacy of sensor data. In addition to privacy concerns, modern large-scale machine learning models can cause environmental impacts resulting from high carbon footprints~\citep{strubell2019energy}. An important direction is to build tiny and efficient models for \iot. The models and resources we have released can help future work explore these efficient methods and quickly benchmark their performance. Finally, we also acknowledge that there is a risk of social biases when human-centric data and possibly sensitive labels are involved, such as models that perform poorly on certain demographic groups. When language model outputs are involved, they can also amplify the underlying social biases~\citep{DBLP:journals/corr/abs-1809-07842} and generate harmful content~\citep{liang2021towards}. Future work must study and mitigate social biases present in multisensory models and large language models.

\newpage

{\footnotesize
\bibliography{reference}
\bibliographystyle{plainnat}
}

\newpage

\appendix
\section*{Appendix}

\section{Details Regarding \iot\ Datasets}

\vspace{-2mm}
\subsection{Twelve Rich Modalities}
\vspace{-1mm}

We collected diverse data from IoT devices, such as Inertial Measurement Units (IMU), Thermal sensors, and Global Positioning Systems (GPS).
Furthermore, we include challenging modalities, such as capacitance, depth, gaze, and pose.
Finally, we collect common and widely used image, audio, and video modalities.
These modalities bring unique challenges since they typically involve noisy real-world sensor measurements, that lack explicit tokenization and alignment with other modalities that we typically expect from conventional multimodal image-text research.

\textbf{IMU:} Inertial Measurement Units capture 3D motion and orientation. This data is fundamental for various applications, including motion tracking and navigation.
We collected 2,940 IMU samples from EyeMU~\citep{kong2021EyeMU} for gaze estimation and motion gesture classification, where they used the accelerometer and gyroscope raw values sampled at 60 Hz as the IMU values per-axis.
28,400 IMU instances are included from SAMoSA~\citep{Mollyn2022samosa} to save synchronized streams of the 9-axis IMU data (accelerometer, gyroscope and orientation) at 50 Hz by using a Fossil Gen 5 smartwatch running Google Android wearOS 2.23.
Further, we sampled 160,120 IMU samples (9-axis) recorded by the device motion sensor using an iOS application~\cite {Riku2022RGBDGaze} on Apple iPhone X for gaze tracking. 
For human bodies, 330,178 IMU orientation recordings~\citep{Huang2018deep} from 17 sensors on different body parts are saved for pose estimation and activity recognition.
For first-person videos in Ego4D~\citep{Grauman2022Ego4D}, we used 510,142 timestamps-based IMU samples with the normalized accelerometer and gyroscope values in each video for activity recognition.
For IMU data on self-driving cars, we collected 41,000 samples from KITTI~\citep{geiger2013kitti} for depth estimation.

\textbf{Thermal:} Thermal modality data provide temperature radiance insights, crucial in surveillance.
For collection, we used 12,025 Thermal samples from LLVIP~\citep{jia2021llvip} containing many pedestrians and cyclists from different locations on the street between 6 and 10 o’clock in the evening. 
They used HIKVISION DS-2TD8166BJZFY-75H2F/V2 as the camera equipment, a binocular camera platform consisting of an infrared camera with a wavelength of $8\sim 14$um.

\textbf{GPS:} Global Positioning Systems offer location data with high precision. This data is invaluable for tasks like location-based services, asset tracking, and navigation.
For GPS data on self-driving cars, we collected 41,000 samples from KITTI~\citep{geiger2013kitti} using OXTS RT3003 inertial and GPS navigation system for depth estimation.
The geographic coordinates include global orientation, altitude, velocities, accelerations, angular rates, and satellite information. 
Following the original dataset, we applied two specified 3-axis coordinates as accelerations for the vehicle and angular rates to describe the tangent plane of the earth's surface corresponding to the geographic location.

\textbf{Camera:} Cameras provide visual data, capturing the environment in rich detail. They serve as the backbone for countless computer vision tasks.
For Camera data, we collected 41,000 instances from KITTI~\citep{geiger2013kitti} using a Velodyne laser scanner installed on a vehicle car for depth estimation.
We stored its 3-axis coordinate and an additional reflectance value for each point. 
The timestamp-based points can be considered according to the scanner's continuous rotation on its vertical axis, which provides a complementary context to GPS/IMU systems for auto-driving.

\textbf{Capacitance:} Capacitive sensors measure changes in capacitance to detect nearby objects or changes. This is foundational for touchscreen technologies and proximity sensing.
For capacitance data, we used 65,374 samples from TouchPose~\citep{Ahuja2021TouchPose} using a 39.6 cm capacitive Crystal Touch panel (Ocular Touch, Dallas, TX), 16-bit touch digitizer, and cameras to record ground-truth data.
When fingers approach the lines on the mutual-capacitance touch sensor, it causes a capacitance drop between lines, resulting in the mutual-capacitance image.

\textbf{Depth:} Depth sensors measure distances between the sensor and objects, providing a 3D view of the environment. They play a significant role in tasks like object detection and scene reconstruction.
For depth data, we collected 160,120
samples from RGBGaze~\citep{Riku2022RGBDGaze} using Apple iPhone X with a TrueDepth camera (640 × 480 depth map interpolated from a 170 × 170 IR dot pattern). 
The participants were asked to look at a target (red dot) that was moving on the screen. 
While the user gazed at the target, the depth imagery was logged at approximately 8 Hz, along with the ARKit gaze prediction. 
For touch depth data, we used 65,374 samples from TouchPose~\citep{Ahuja2021TouchPose} recorded by a ToF depth camera (Azure Kinect) below the surface of the transparent touch panel. 
The depth modality is essential for touch contact and 3D pose joint reconstruction.

\textbf{Gaze:} Gaze sensors track eye movement and direction, offering insights into user attention and intention.
Regarding gaze modality, we collected 2,940 samples from EyeMU~\citep{kong2021EyeMU} by running an iOS application on an Apple iPhone 12 Pro (screen size is 12.8 × 6.4 cm).
The participants were asked to gaze at a single red dot, and the screen advanced to capture a motion gesture and a 2-axis gaze location after 1.2 seconds. 
Furthermore, we used 160,120 gaze samples from RGBGaze~\citep{Riku2022RGBDGaze} by running an iOS application on Apple iPhone X to record gaze tracking data with pre-determined 35 fixed locations on the screen.

\textbf{Pose:} 
Pose sensors capture the orientation and position of objects or individuals, which is critical for motion analysis and interactive applications.
For body pose data, we collected 330,178 samples from DIP-IMU~\citep{Huang2018deep} using  Xsens IMU sensors containing 3-axis accelerometers, gyroscopes, and magnetometers.
They placed the head sensor onto the head of each participant such that the sensor axes aligned with the SMPL body frame to do a calibration.
The SMPL pose parameters are stored in the angle-axis format with three joint angles (yaw, pitch, roll) per 24 joints.
For touch pose data, we used 65,374 samples in TouchPose~\citep{Ahuja2021TouchPose} from a Leap Motion stereo IR camera, running Orion 4.1 for 3D hand pose tracking.  
We included 14 different finger and whole-hand touch poses and gestures, representing each pose with 3030 to 5875 samples.

\textbf{LiDAR:} LiDAR sensors emit light to measure distances, generating high-resolution 3D maps of environments. They are central to applications like autonomous driving and topographical mapping.
For LiDAR data, we collected 51,000
samples from the Newer College dataset~\citep{Ramezani2020newer} using the Ouster LiDAR with 64 beams, 64 Channels, 120 m range, 45$^\circ$ vertical Field-of-View (FoV), and 1024 horizontal resolution.
During the collection, the Ouster LiDAR synchronized with the recording computer using the Precision Time Protocol (PTP) to achieve sub-microsecond accuracy.
The accurate prior map of 290 million points was down-sampled to 1 cm resolution and reduced to about 17 million points, allowing for its use without an observable drop in registration accuracy. Further, cropping the reduced point cloud around the sensor's pose dynamically created the final reference cloud in 100 m by 100 m.

\textbf{Video:} Video captures sequences of visual frames, providing a dynamic view of the environment. 
This modality supports tasks ranging from action recognition to anomaly detection.
For video modality, we used 510,142
egocentric videos with 30FPS in Ego4D~\citep{Grauman2022Ego4D}, which includes a wide range of everyday activities, such as cooking, cleaning, and fishing.
These videos also cover diverse geographic locations across the world, and are paired with timestamps-based
IMU values of the normalized accelerometer and gyroscopes.

\textbf{Audio:} Audio sensors capture sound waves, enabling voice recognition, sound classification, and environmental sound analysis.
For audio data, we collected 28,400 samples paired with IMU modality from SAMoSA~\citep{Mollyn2022samosa}, where participants wore the smartwatch on their dominant arm, and were asked to perform 26 activities across 4 contexts with each activity repeated 3 times within each context.
As the audio was sampled at 1 kHz, the resolution of the information went down, and more activity classes, such as Hand Washing and Toothbrushing, got similar and confused. 
In such cases, IMU data can provide valuable information to remove ambiguity.

\textbf{Image:} Image sensors offer static visual captures of the environment, serving as a basis for a myriad of vision tasks.
For RGB image data, we collected 160,120 samples from RGBDGaze~\citep{Riku2022RGBDGaze} paired with gaze, depth, and IMU for gaze tracking.
To align GPS and Camera modalities with images, we collected 41,000 samples from KITTI~\citep{geiger2013kitti} for depth estimation and activity recognition.
Furthermore, we used 12,025 high-quality images paired with infrared thermal samples in LLVIP~\citep{jia2021llvip} from 26 locations.
For alignment with body pose, we used 330,178 samples from DIP-IMU~\citep{Huang2018deep} for pose estimation and activity recognition.
Regarding hand pose images, we collected 65,374 instances from TouchPose~\citep{Ahuja2021TouchPose} for touch contact classification and 3D hand pose joint reconstruction.

\vspace{-2mm}
\subsection{Eight Well-defined and Challenging Tasks}
\vspace{-1mm}

Our benchmark includes tasks that reflect real-world IoT challenges and that will drive the community towards solutions with tangible societal impacts.

\textbf{Gaze estimation:} This task is pivotal for human-computer interaction, driver monitoring, and virtual reality.
Given RGB images of faces, depth and IMUs, our goal is to predict the location (X/Y) for tracking gazes of the person. 
This regression task requires multisensory understanding on long-range interactions between RGB images and depth and heterogeneity in IMUs.

\textbf{Depth estimation:} A cornerstone for AR/VR applications, robotics, and object detection, depth estimation involves predicting the distance between the camera and each pixel in the image.
Given RGB images, camera parameters, GPS coordinates, and IMU, we are expected to predict the depth maps of objects, such as cars and pedestrian on the streets. 
In the touch robots case, given RGB images, capacitive image, and hand poses, our target is to estimate the depth maps of hands. 
This regression problem requires multisensory understanding on long-range interactions between RGB images and capacitance and heterogeneity in poses.

\textbf{Gesture classification:} Crucial for intuitive human-machine interfaces, gesture classification aims to recognize specific hand or body movements.
Given gaze locations and IMU data on accelerometer, gyroscope and orientation, the task is defined to classify the gesture of human heads.
This classification problem requires the cross-model perception on heterogeneity in gaze and IMUs.

\textbf{Pose estimation:} With applications in AR/VR, gaming, and health, pose estimation focuses on determining the spatial arrangement of human joints.
Given RGB images and measured IMU data, our goal is to predict the poses of human body including 24 joints with three joint angles (yaw, pitch, roll). 
This regression problem requires a deeper cross-modal understanding on the heterogeneity in IMUs and RGB pixels.

\textbf{Touch contact classification:} Vital for enhancing user experiences on touch-based devices, this task involves determining the type or nature of touch on capacitive surfaces.
Given RGB images, capacitive images, depth maps, and hand poses, we are expected to classify touch contact using diverse modalities.
This classification task requires a multimodal understanding on the long-range interactions between RGB images and capacitance and heterogeneity in depth maps and poses.

\textbf{Event detection:} A broad area with applications in surveillance, smart homes, and industrial setups, event detection involves identifying specific occurrences or anomalies in the data stream.
Given audio spectrograms and  IMU data on accelerometer, gyroscope and orientation, our goal is to predict the categories of events across different timestamps.
This classification problem requires a cross-modal understanding on the long-range interactions between audio and IMU.
If a predicted activity is above a confidence threshold, we consider it an event. 
Othwise, if it’s below a confidence threshold, or belongs to the Other class, we do not consider it an event.

\textbf{Activity recognition:} Central to fitness, health, and elder care applications, activity recognition aims to discern human activities like walking, running, or jumping.
Given RGB images, poses with three joint angles (yaw, pitch, roll), and IMU data, we are expected to classify the class of actions for the human body.
For ego-centric cases, we are given video frames and IMU orientation recordings on from different sensors to predict the category of activity in the videos.
This classification task requires a cross-modal understanding on the heterogeneity in poses, videos and IMU.

\textbf{3D reconstruction:} With significance in gaming, film, and AR/VR, 3D reconstruction involves creating a three-dimensional model of an environment or object from 2D data.
Given RGB images, capacitance image, and depth maps, our target is to reconstruct the 3D poses. 
This regression problem requires a multimodal understanding of both capacitance images and depth maps.

\section{Experimental Setup}

\subsection{Setup for Unimodal Models}
\begin{itemize}
    \item \textbf{Data Preparation:} Each modality, e.g., RGB images, capacitive images, or hand pose, is pre-processed independently. The data undergo normalization and any specific transformations tailored to that modality.
    \item \textbf{Network Architecture:} Distinct neural architectures optimized for each modality type, such as CNNs for images and RNNs for sequential data.
    \item \textbf{Training Details:} Models are trained using a batch size of 128, employing the Adam optimizer with a learning rate of 0.001. Early stopping with a patience of 10 epochs ensures prevention from overfitting.
    \item \textbf{Evaluation:} Each unimodal model is evaluated on its respective validation dataset to gauge performance.
\end{itemize}

\subsection{Setup for Adapter Models}
\begin{itemize}
    \item \textbf{Data Preparation:} Data is fed through a pre-trained network, where only the adapter modules are trainable.
    \item \textbf{Network Architecture:} Utilizing deep architectures like LLaMA~\citep{gao2023llamaadapterv2}, but with adapter layers inserted in-between the pre-defined layers.
    \item \textbf{Training Details:} Since only the adapter layers are trainable, fewer parameters are updated, allowing for a larger batch size of 256. The training uses the Adam optimizer with a learning rate of 0.0005.
    \item \textbf{Evaluation:} Model performance is assessed by evaluating the fine-tuned model on the targeted task’s validation set.
\end{itemize}

\subsection{Setup for Unimodal Multi-task Models}
\begin{itemize}
    \item \textbf{Data Preparation:} Data from different tasks, but the same modality, are concatenated or paired.
    \item \textbf{Network Architecture:} Shared encoder layers process the input data, followed by task-specific decoders.
    \item \textbf{Training Details:} Gradient balancing techniques are employed to prevent one task from dominating the training process. Training leverages a batch size of 128 and the Adam optimizer with a learning rate of 0.001.
    \item \textbf{Evaluation:} Performance is evaluated separately for each task on their respective validation sets.
\end{itemize}

\subsection{Setup for Multisensory Models}
\begin{itemize}
    \item \textbf{Data Preparation:} Data from different modalities are fused either at the input, feature, or decision level.
    \item \textbf{Network Architecture:} Modality-specific encoders process each input type. Fusion layers then combine features from all encoders.
    \item \textbf{Training Details:} Models are trained with a batch size of 128 using the Adam optimizer and a learning rate of 0.001. Data balancing techniques ensure equal representation from each modality.
    \item \textbf{Evaluation:} The combined model's efficacy is evaluated using a validation dataset that includes all modalities.
\end{itemize}

\subsection{Setup for Multisensory Multitask Models}
\begin{itemize}
    \item \textbf{Data Preparation:} Data from different modalities and tasks are paired or concatenated as required.
    \item \textbf{Network Architecture:} Shared modality-specific encoders are followed by task-specific decoders.
    \item \textbf{Training Details:} Gradient balancing techniques are applied, along with modality balancing, to ensure fairness in learning. The model trains using a batch size of 128 and the Adam optimizer at a learning rate of 0.001.
    \item \textbf{Evaluation:} Each task's performance is assessed on their respective validation datasets.
\end{itemize}

\noindent For all the methods, the experimental environment remains consistent. All models are trained and evaluated on NVIDIA V100 GPUs, ensuring uniformity in computational power and performance.

\section{Evaluation Metrics}

To measure performance, we utilize a combination of metrics following prior work on each specific task.
For gaze estimation, we use mean euclidean error in centimeters to measure the positional distance between the predicted gaze and the ground-truth gaze.
For depth estimation, we apply mean absolute error in millimeter to calculate the gap between the prediction and the ground-truth depth.
For gesture classification, we compute the ratio of correct classified samples as the accuracy.
For pose estimation, we use mean euclidean error in centimeters to measure the positional distance between the predicted pose joints and the ground-truth pose.
For touch contact classification, we calculate the accuracy of classifying the category of fingers interacting with the touchscreen.
For event detection, we apply F1 score to decide if the predicted activity above a confident threshold belongs to a event.
For activity recognition, we compute the balanced accuracy for measuring instance-level performance.
For 3D pose reconstruction, we use End-point-error in millimeter, the mean Euclidean error between all the joints of the annotated and predicted hand pose.

\section{More analysis}

\textbf{Testing long-range interactions}: Long-range interactions are critical to many problems in machine learning, particularly in fields like time series forecasting, natural language processing, and signal analysis. 
Recognizing patterns and relationships over vast sequences or across multiple modalities often requires models to understand and leverage these long-range dependencies. 
However, capturing these interactions remains a challenge for many conventional models.

In a controlled experiment, we truncated sequences to various lengths and observed how conventional models performed. 
As the sequence lengths increased, representing longer durations of time or more extensive contexts, there was a marked decline in performance. 
This showcased the models' inability to effectively encapsulate and understand interactions beyond a certain range.
Multimodal setups further complicate this. 
The long-range dependencies aren't just within a modality but can also be across modalities. 
This inter-modality long-range interaction is a largely uncharted territory, and our experiments showed that it's an area where even advanced models can falter.

Exploring architectures that inherently focus on long-range interactions, potentially leveraging self-attention mechanisms but with modifications to handle extremely long sequences.
Employing models that operate at different temporal scales, allowing them to summarize information at various levels and potentially capture longer-range interactions more effectively.
Techniques that allow models to allocate more computational resources when faced with potential long-range dependencies, thus emphasizing critical parts of a sequence or modality.
For multimodal problems, mechanisms that facilitate better cross-modal attention can be crucial. 
This will enable models to recognize and act upon dependencies that span across different modalities, even if they are separated by considerable temporal or sequential gaps.

\textbf{Testing heterogeneity in structure and noise}: Heterogeneity in data, both in terms of structure and noise, is a pervasive challenge in machine learning. 
As datasets grow more complex, encompassing a wider variety of sources, the inherent differences in data structure and the presence of various types of noise can significantly hamper the performance of models. 
Understanding how models grapple with such heterogeneity is vital for real-world applications.

We exposed models to datasets that combined structured data (such as GPS, IMU) with unstructured data (such as images or raw audio). 
Unimodal baselines often struggled to reconcile these different data forms, leading to a significant drop in accuracy compared to when dealing with homogenous data types.
We also introduced varying degrees of noise into datasets, Gaussian noise in numerical data. 
Currrent methods saw a rapid decline in performance as the noise levels increased, unable to filter out irrelevant information effectively.
Heterogeneity challenges underline the importance of robustness in model design. 
Our experiments highlighted that many models, even those considered state-of-the-art, have vulnerabilities when exposed to unexpected data structures or noise patterns.

Exploring architectures and training techniques that are inherently more robust to noise and heterogeneity. This might include noise injection during training or techniques like dropout that encourage model generalization.
Leveraging advanced data augmentation techniques, both for structured and unstructured data, to simulate and thus prepare the model for varied data structures and noise patterns.
Using meta-learning approaches where models are trained to quickly adapt to new data structures or noise patterns with minimal fine-tuning.
Research into advanced denoising mechanisms, especially ones that can handle structured noise, can be invaluable. 
This includes both pre-processing methods and in-model techniques.

\vspace{-2mm}
\subsection{Analysis of information sharing}
\vspace{-1mm}


Finally, we show examples of how information is shared across modalities and tasks, based on two potential sources of sharing: low-level modality features and high-level semantic concepts.

\textbf{Low-level modality features}:
Different sensory modalities often contain unique low-level perceptual features that complement those in other modalities. We illustrate this information sharing across 3 modalities: IMU, video, and pose data for predicting 2 common activities: walking and dancing.

\textit{Walking} is a common activity with distinctive rhythmic characteristics. Using IMU features, the model learns that rhythmic patterns, particularly in acceleration and deceleration, correspond to each walking step. The cadence, stability, and any irregularities in the walking pattern can also be inferred. Video features capture the holistic visual representation of walking, presenting details such as gait, arm swing, speed, stride length, and frequency. Finally, pose features highlight the specific posture changes during walking, emphasizing leg movement, foot placement, and body alignment.

\textit{Dancing} requires complex and expressive motions with varying styles and dynamics. IMU provides dynamic, often non-linear patterns in IMU data, reflecting the dance's tempo, vigor, and style variations; video captures the dance form, style, synchronization, and expressiveness; and pose data captures the alignment and configuration of body parts, offering insights into dance postures, transitions, and intricate footwork or hand movements.

\textbf{High-level semantic concepts}
encapsulate a more general conceptual understanding and reasoning about the environment. We show two examples showing how the audio and IMU modalities share information about two high-level semantic concepts, focusing on 'body pose' and 'hand pose'.

\textit{Body pose} represents the spatial arrangement and posture of the entire human body. This can involve stances like standing, sitting, lying down, or even dynamic movements like jumping or running. For Audio, indirect cues such as the sound of footsteps, a person sitting down on a chair, or even the echo in a room (indicating a certain body pose affecting sound propagation) can provide hints about the body's posture.
For IMU, accelerometers capture the directional movement while gyroscopes provide rotational dynamics to distinguish if a person is upright, moving rapidly, or stationary.

\textit{Hand pose} looks at the orientation, gesture, and spatial arrangement of just the hands, ranging from gestures like waving, gripping, to more intricate signs in sign language. In audio, sounds like clapping, snapping, or even the subtle rustling of hands moving through the air can be detected. The distinct sounds made by hang interactions with objects can also hint at specific hand poses. When IMU sensors are placed on the wrist or back of the hand, they can capture detailed dynamics of hand movements, tilting, rotation, or swift movements that indicate hand poses.

\section{More examples}

In this section, we present more examples of diverse IoT modalities, emphasizing their heterogeneity and the implications of their temporal interactions. 
Each modality contributes uniquely to the understanding and processing of environmental data, pivotal for applications in IoT networks.
These examples underscore the heterogeneity in sensor types and data characteristics in IoT systems. 
Moreover, they highlight the importance of temporal interaction in data processing and application responsiveness.

\begin{figure}[t]
\centering
\includegraphics[width=0.98\linewidth]{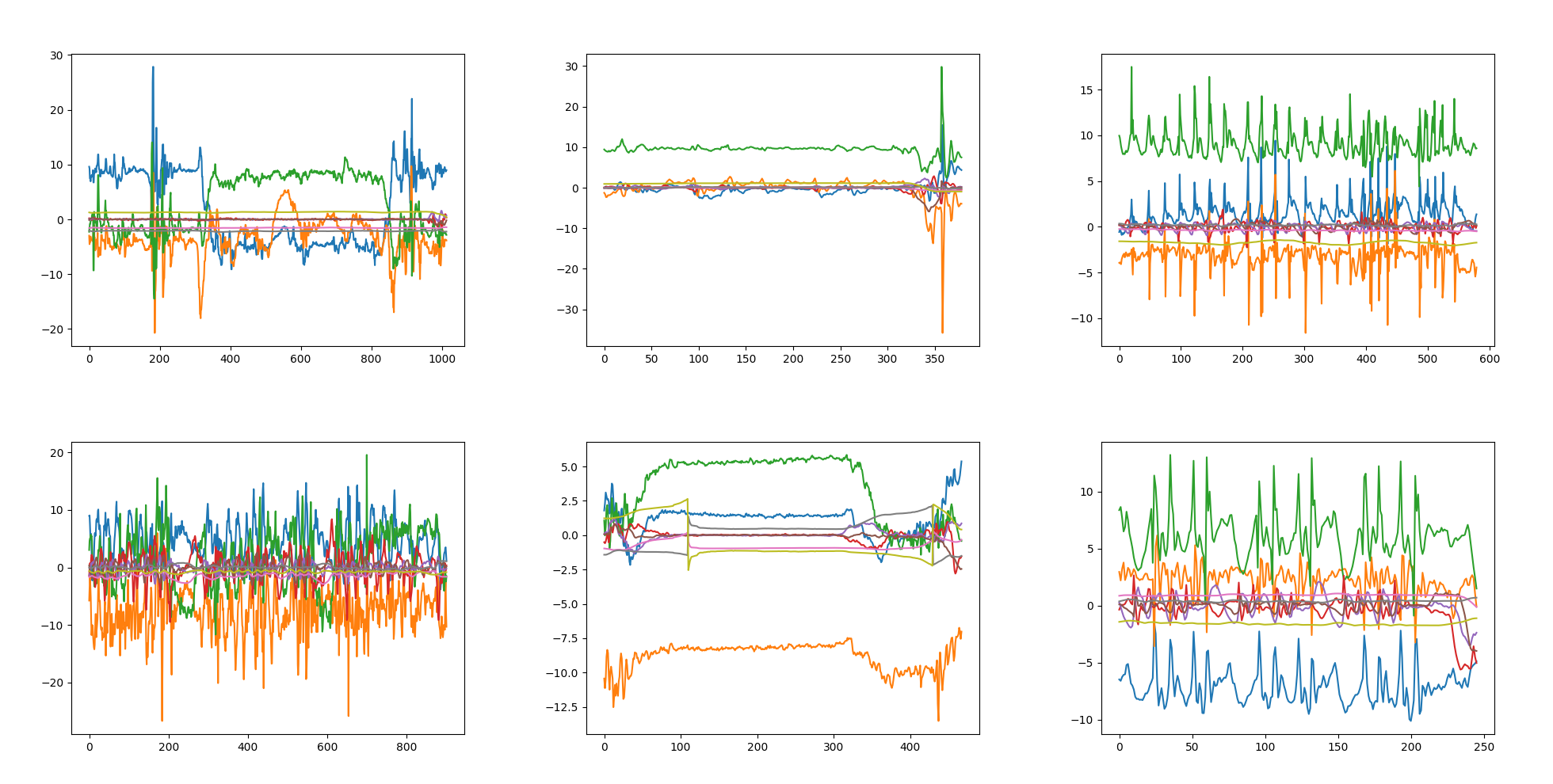}
\vspace{-4mm}
\caption{IMU Visualizations}
\vspace{-4mm}
\label{fig: vis_imu}
\end{figure}

\subsection{IMU}

The Inertial Measurement Unit (IMU) is critical for capturing dynamic motion and orientation. 
An IMU typically combines accelerometers, gyroscopes, and sometimes magnetometers to provide comprehensive motion tracking. 
For example, in a smartwatch, the IMU captures temporal data on user movement patterns, crucial for activity recognition and health monitoring applications. 
This modality's high sampling rate allows for detailed temporal analysis, capturing minute fluctuations in motion, as shown in Figure~\ref{fig: vis_imu}.

\begin{figure}[t]
\centering
\includegraphics[width=0.98\linewidth]{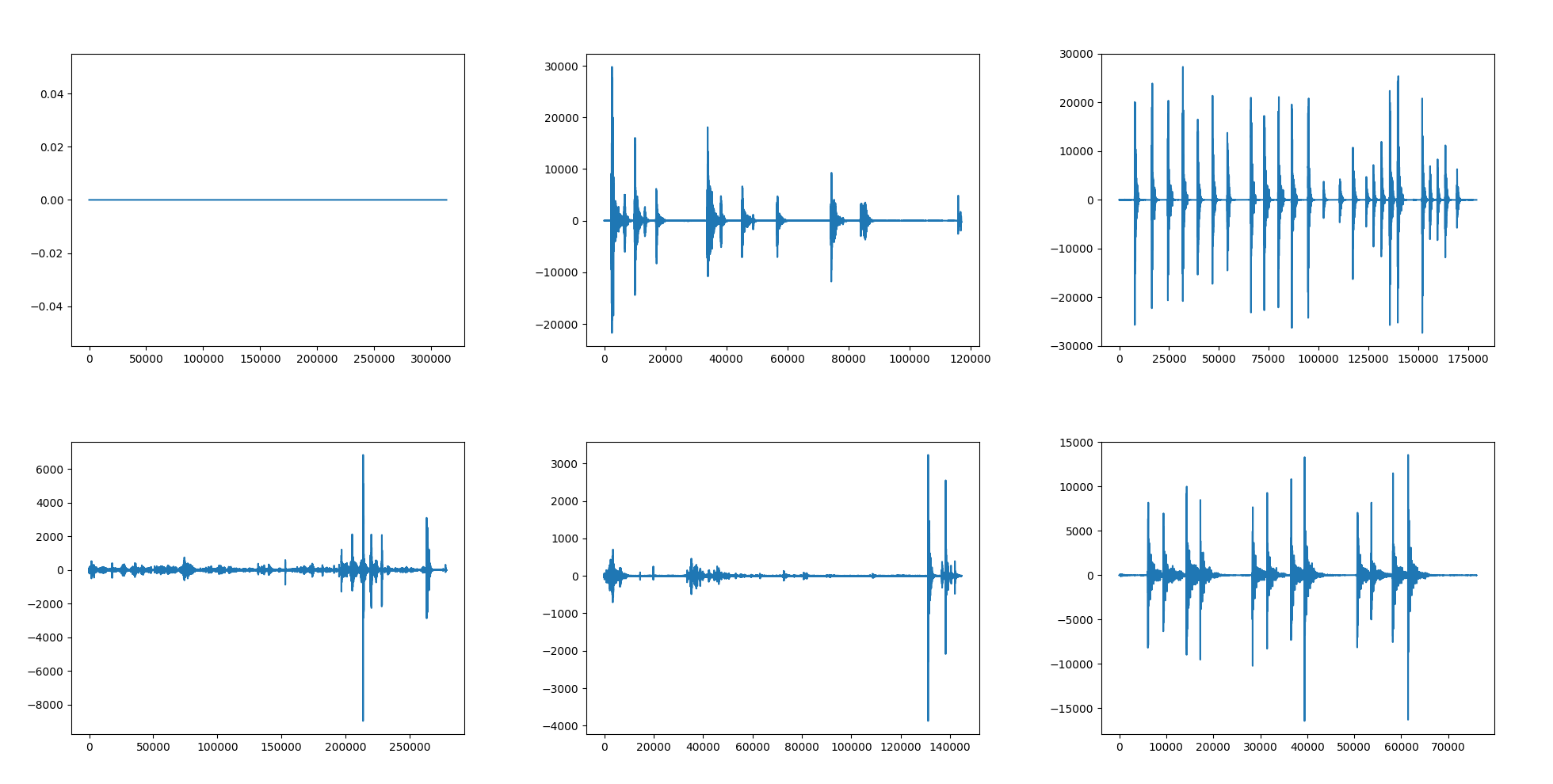}
\vspace{-4mm}
\caption{Audio Visualizations}
\vspace{-4mm}
\label{fig: vis_audio}
\end{figure}

\subsection{Audio}

Audio sensors capture sound waves, converting them into digital signals that represent the acoustic environment. 
In smart homes, audio sensors can detect various sounds, from spoken commands to the activity noise of household appliances. 
The temporal granularity of audio data reported in Figure~\ref{fig: vis_audio} is vital for applications like speech recognition, environmental sound classification, and emergency detection (e.g., breaking glass or alarms).

\begin{figure}[t]
\centering
\includegraphics[width=0.98\linewidth]{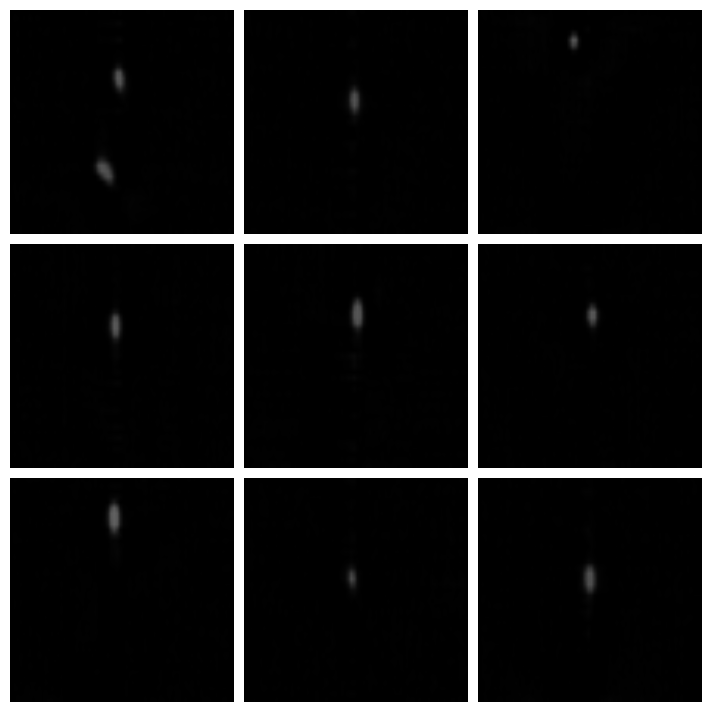}
\vspace{-4mm}
\caption{Capacitance Visualizations}
\vspace{-4mm}
\label{fig: vis_cap}
\end{figure}

\subsection{Capacitance}

Capacitive sensing involves the measurement of changes in capacitance in an environment, often used to detect touch or proximity. 
In an IoT context, capacitive sensors can be embedded in surfaces to create interactive touch interfaces or to monitor object presence and human interaction without direct contact. 
As shown in Figure~\ref{fig: vis_cap}, the temporal resolution of capacitance can vary, but its real-time response is crucial for interactive applications.

\begin{figure}[t]
\centering
\includegraphics[width=0.98\linewidth]{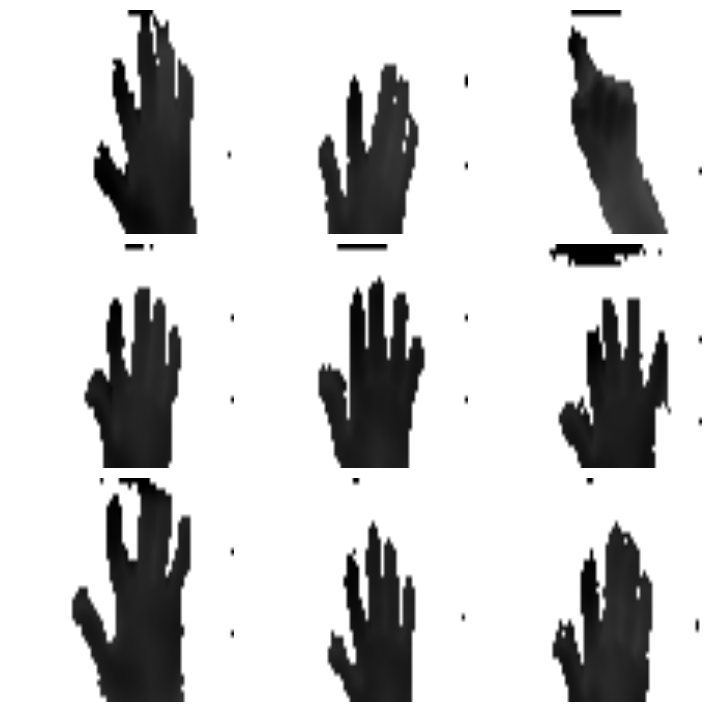}
\vspace{-4mm}
\caption{Depth Visualizations}
\vspace{-4mm}
\label{fig: vis_depth}
\end{figure}

\subsection{Depth}

Depth sensors measure the distance between the sensor and objects in its environment, typically using technologies such as LIDAR, structured light, or time-of-flight cameras.
This modality is essential in scenarios where spatial relationships and object recognition are required, such as in autonomous vehicle navigation or interactive gaming, as illustrated in Figure~\ref{fig: vis_depth}.
Temporal interactions in depth sensing are crucial for understanding scene changes and movement dynamics over time.


\end{document}